\documentclass[runningheads]{llncs}

\usepackage{eccv}

\usepackage{eccvabbrv}

\usepackage{graphicx}
\usepackage{booktabs}

\usepackage[accsupp]{axessibility}

\usepackage{amsmath,amssymb,amsfonts}

\usepackage{hyperref}

\usepackage{orcidlink}

\newcommand{\Magrini}{Magrini \etal \cite{Magrini2025-FRED} }
\newcommand{\HariPrasanth}{Hari Prasanth \etal \cite{HariPrasanth2026} }

\newcommand{\state}{\mathbf{x}}

\newcommand{\repr}{\mathcal{R}}

\newcommand{\bbox}{\mathbf{b}}

\usepackage[absolute]{textpos}
\definecolor{somegray}{gray}{0.6}
\newcommand{\darkgrayed}[1]{\textcolor{somegray}{#1}}

\begin{document}

\begin{textblock}{8}(4, 0.8)
\begin{center}
\darkgrayed{Accepted to the Workshop on Neuromorphic Vision (NeVi) of the European Conference on Computer Vision (ECCV), 2026.}
\end{center}
\end{textblock}

\title{Residual Kalman Dynamics for Event-Based UAV Forecasting} 

\author{Per Nyblom\orcidlink{0009-0005-9828-2548} \and
Hannes Ovrén\orcidlink{0000-0003-3365-4062} \and
David Gustafsson\orcidlink{0000-0002-4370-2286}}

\authorrunning{P.~Nyblom et al.}

\institute{Swedish Defence Research Agency (FOI), Linköping, Sweden\\
\email{\{per.nyblom, hannes.ovren, david.gustafsson\}@foi.se}\\
}

\maketitle

\begin{abstract}

We study short- and mid-horizon UAV bounding-box forecasting on the FRED event-camera dataset. We use a constant-velocity Kalman filter over a full center-size box state as a strong physical baseline, and train a residual model to predict acceleration-like corrections from recent box history, filtered state features, and local event representations. This simple residual formulation consistently improves over the Kalman baseline, with event-conditioned models giving the strongest results among the evaluated methods. We further show that part of the residual target is predictable from anchor position and velocity alone, indicating that canonical FRED results can reflect both visual evidence and dataset-specific motion priors. To analyze this effect, we introduce decorrelated subsets as a diagnostic stress test, showing that event-conditioned residual models retain useful predictive signal even when measured position- and velocity-based shortcuts are weakened.
 
\end{abstract}

\section{Introduction}
\label{sec:intro}

Unmanned aerial vehicles (UAVs) are increasingly used in civilian, industrial, and security-critical settings, making reliable perception of small, fast-moving aerial objects important for detection, tracking, and motion anticipation ~\cite{sivakumar2021uavsurvey}. Recent incidents and conflicts have further underscored the need for robust UAV and counter-UAV perception systems~\cite{bbc2019,bbc2025,ModUA2026}. Short-horizon UAV trajectory forecasting is therefore relevant for systems that must anticipate rapid aerial maneuvers under low latency.

Using traditional RGB cameras for UAV trajectory forecasting\cite{Becker2021,Zou2025,Zhou2024} is problematic due to the high agility of the targets together with the relatively low frame rate of conventional cameras.
Event cameras\cite{Gallego2022} are a compelling alternative and have previously been used with success for \eg UAV tracking \cite{Magrini2025-DroneDet}.
Unlike traditional cameras, event cameras detect changes in light intensity and output these asynchronously at microsecond temporal resolution with low latency and high dynamic range.
This means that an event camera can detect and track UAVs under low contrast, during aggressive maneuvers, and even measure the rotation speed of their propellers\cite{HariPrasanth2026,Stewart2023,Spetlik2025}.

Forecasting UAV trajectories using event cameras is, however, a relatively new research direction.
Liang \etal \cite{Liang2025} developed a method that fused RGB images with events,
however, their events were simulated from low frame rate RGB images that cannot represent the temporal characteristics of a UAV observed by a real event camera.
To the best of our knowledge, there are only two examples of previous work that use real event camera data for UAV trajectory forecasting \cite{Magrini2025-FRED,HariPrasanth2026}, and we choose to focus on these works here.
For an overview on tracking and detection of drones using event cameras we instead refer the reader to a recently published survey article\cite{Magrini2025-DroneDet}.

\begin{figure}[tb]
    \centering

    \begin{subfigure}{0.48\textwidth}
        \centering
        \includegraphics[width=\linewidth]{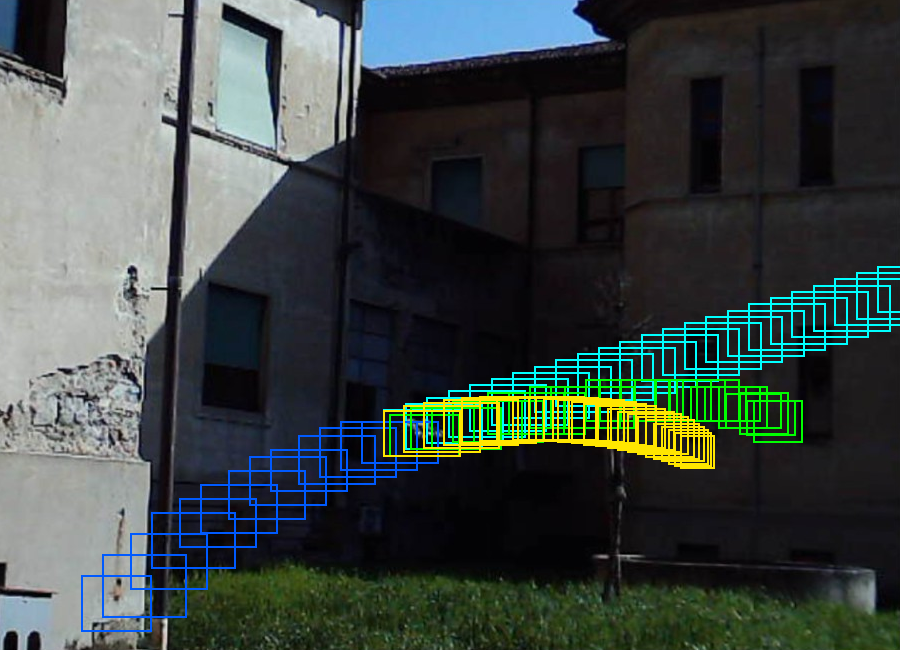}
    \end{subfigure}
    \hfill
    \begin{subfigure}{0.48\textwidth}
        \centering
        \includegraphics[width=\linewidth]{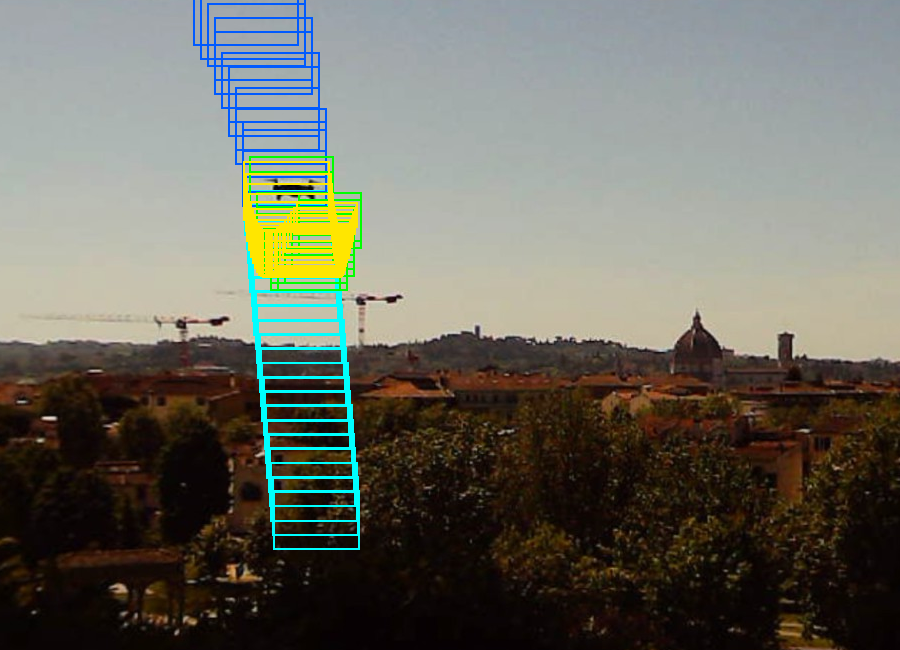}
    \end{subfigure}

    \vspace{0.2cm}

    \begin{subfigure}{0.48\textwidth}
        \centering
        \includegraphics[width=\linewidth]{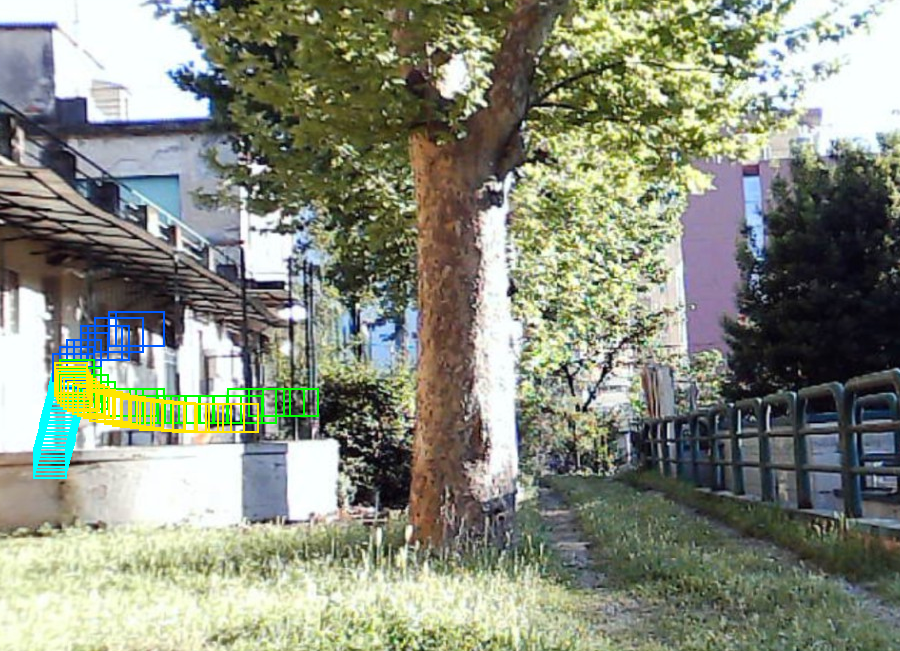}
    \end{subfigure}
    \hfill
    \begin{subfigure}{0.48\textwidth}
        \centering
        \includegraphics[width=\linewidth]{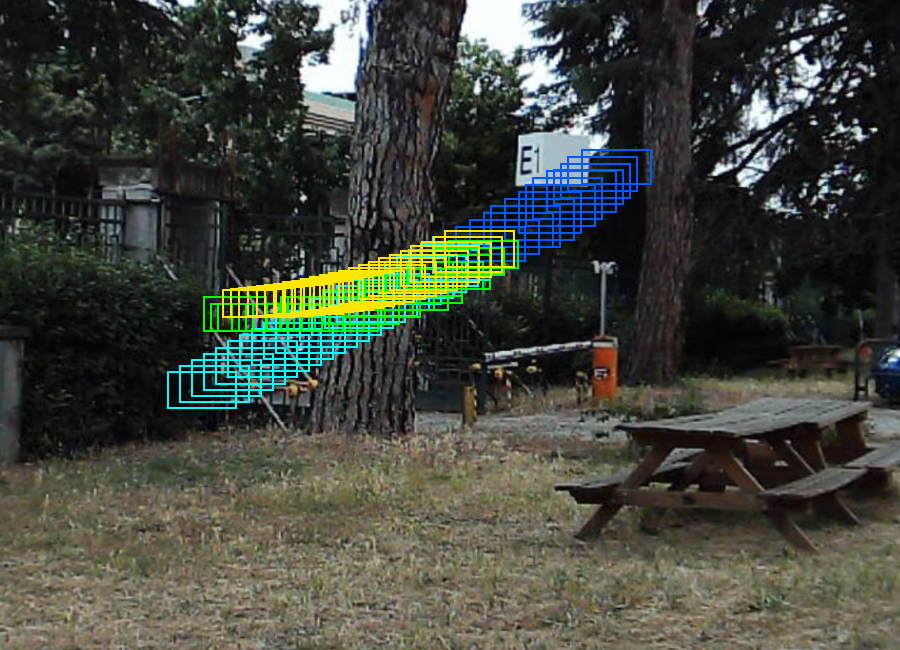}
    \end{subfigure}

    \caption{Qualitative examples from the FRED dataset showing the effect of acceleration-aware residual forecasting. Observed history boxes are shown in blue, future ground-truth boxes in green, constant-velocity Kalman predictions in cyan, and residual-model predictions in yellow. The constant-velocity model extrapolates the recent motion direction and can deviate substantially when the UAV accelerates or turns, whereas the residual model better follows the future trajectory by learning acceleration-like corrections to the Kalman prediction. }
    \label{fig:forecasting_examples}
\end{figure}

An enabling factor for event-based UAV trajectory forecasting was the release of \emph{The Florence RGB-Event Drone Dataset (FRED)}\cite{Magrini2025-FRED} in 2025, which contains seven hours of recorded UAV flights with different platforms. 
The dataset is annotated with bounding boxes, and is geared towards UAV detection, tracking, and trajectory forecasting.
As part of their work on the FRED dataset, \Magrini also developed and benchmarked five baseline forecasting methods with different architectures and input modalities (RGB images, events, or both).
In their experiments they show that using events gives a slight edge over using RGB images,
however, the improvement is marginal compared to the Transformer that is trained on only the bounding boxes.

\HariPrasanth propose a method that does not involve any machine learning.
Instead, they use a Kalman filter with a constant velocity motion model which is updated using position and velocity measurements extracted from the previous bounding box centers, and the forecasted trajectory is predicted by the filter (omitting measurements).
They incorporate event camera data by using it to estimate the RPM of the UAV propellers, and then use this to modulate the process noise covariance matrix of the Kalman filter.
The intuition is that a high RPM indicates aggressive flight patterns, which means that the filter update should place less weight on the motion model, and more on the measurements.
Evaluation on the FRED dataset shows a significant improvement over the previous state of the art\cite{Magrini2025-FRED},
and that the addition of event-based RPM-modulated process noise gives a small improvement over a standard Kalman filter.
Interestingly, they show that simple linear extrapolation of bounding-box centers is almost as good as using the Kalman-filter approach.

Following \HariPrasanth, we use a constant-velocity Kalman model as a simple and directly comparable physical prior. 
Rather than increasing the order of the dynamics alone, we learn residual acceleration corrections from box history, filtered state features, and event evidence.
In many applications, being able to estimate target size is also important, and because of this we predict full bounding boxes instead of only box centers.
We also do not modulate the process noise using RPM estimation.
Instead, our approach uses the Kalman filter as a reliable physical prior, but improves the prediction by also training a machine-learning model to predict acceleration-like corrections to the predicted filter states. 
This formulation separates a reliable physical prior from the remaining predictable structure in the data.

The benefits of this approach are illustrated in Figure \ref{fig:forecasting_examples}, which shows four FRED sequences where a constant-velocity Kalman filter fails to anticipate changes in the UAV motion. In these examples, the predicted boxes continue along the recent velocity direction, whereas the ground-truth future trajectory bends due to acceleration. 
A residual model that predicts acceleration-like corrections can therefore better follow the future trajectory while retaining the stability of the Kalman prior.

A central difficulty in evaluating short-horizon UAV forecasting is that good performance need not imply that a model has learned to use target appearance or event evidence.
Depending on how the dataset was recorded in terms of \eg camera viewpoint and UAV flight patterns, the future motion of a target can be partially predictable from its current image location and velocity alone. For example, targets near the image boundary may be more likely to turn back toward the center of the field of view, and high-speed trajectories may be more likely to decelerate. Such regularities are useful for in-distribution forecasting, but they can also act as motion-prior shortcuts that obscure whether an event-conditioned model is actually exploiting visual evidence in contrast to simply learning dataset-specific trajectory statistics.

This motivates a diagnostic question: how much of the residual forecasting problem can be explained without looking at the event data? 
We address this by measuring how well estimated future accelerations can be predicted from simple anchor features such as image position and velocity,
and propose a diagnostic that checks whether the same modeling choices remain useful when these simple motion priors are weakened.

To construct this diagnostic, we introduce a sample-level decorrelation procedure that greedily removes forecast windows contributing most to the predictability of future acceleration from anchor position and velocity. We use the resulting decorrelated subsets as a stress test for motion-prior dependence. 
Comparing models before and after this procedure helps identify which gains persist when position- and velocity-based shortcuts are reduced.

In summary, our contributions are
\begin{itemize}
\item a residual Kalman forecasting model (see \cref{sec:residual_kalman}) that predicts acceleration corrections on top of a full center-size box state, improving both short- and mid-horizon UAV bounding-box forecasts on FRED.
\item an empirical diagnostic (see \cref{sec:motion_prior_corr})  showing that future acceleration in FRED is partially predictable from simple motion-prior features such as anchor position and velocity.
\item a greedy sample decorrelation procedure (see \cref{sec:greedy_decorrelation}) used as a stress test for motion-prior dependence, enabling a more cautious analysis of whether learned gains persist when position- and velocity-based shortcuts are weakened.
\end{itemize}

\section{Problem Setup}

\subsection{State and Forecast Target}

Each sample is anchored at label time $t_0$. 
The observed history contains boxes up to and including $t_0$, and the forecasting target is the sequence of future boxes after $t_0$. 
We represent a bounding box 
\begin{equation}
  \mathbf{b}_t = [x_t, y_t, w_t, h_t]^\top \in [0,1]^4,
  \label{eq:kalman_state}
\end{equation}
at time $t$ in normalized center-size coordinates where ($x_t$,$y_t$) is the box center and ($w_t$,$h_t$) are its width and height. All coordinates are normalized by the image width and height.

The Kalman state at time $t$,
\begin{equation}
  \state_t =
  [x_t, y_t, w_t, h_t, v^x_t, v^y_t, v^w_t, v^h_t]^\top,
\end{equation}
augments the box with first-order velocities for all four box channels,

This full-center state allows the baseline and residual models to forecast bounding boxes rather than only center trajectories. 
We use a 400 ms history window and evaluate both short-horizon 400 ms and mid-horizon 800 ms forecasts, following the stated forecast protocol in \Magrini.

\subsection{Inputs}

\begin{figure}[tb]
    \centering

    \begin{subfigure}{0.48\textwidth}
        \centering
        \includegraphics[width=\linewidth]{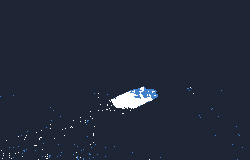}
    \end{subfigure}
    \hfill
    \begin{subfigure}{0.48\textwidth}
        \centering
        \includegraphics[width=\linewidth]{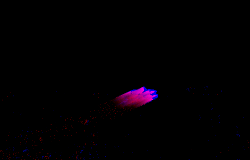}
    \end{subfigure}

    \caption{Event-derived input representations used by the image-conditioned residual models. Left: the event-frame representation provided with FRED. Right: the same event sequence rendered as a compact spatio-temporal representation (CSTR) \cite{cstr}. }
    \label{fig:rep_formats}
\end{figure}

We evaluate three families of input features. First, we use event-derived image representations extracted from the event stream immediately preceding the anchor time, $t_0$. Shown in \cref{fig:rep_formats} are the two selected spatio-temporal representations: CSTR \cite{cstr}, and event frames provided by the FRED dataset. CSTR provides a compact three-channel rendering of recent event activity, while event frames use a per-polarity binary representation. Second, we use recent box history, represented as normalized center-size boxes with relative timestamps. Third, we use features derived from the optimized Kalman filter state, including filtered position and velocity estimates.

To reduce direct leakage of global image position into image-conditioned models, the main experiments use fixed spatial cutouts centered on the final observed box. Inputs are resized to $(640 \times 360)$, and $(64 \times 64)$ cutouts are extracted around the anchor box, corresponding to $(128 \times 128)$ pixels at the original FRED resolution. The history encoder uses relative box-history features in the main setting, where positions are expressed with respect to the final observed box. Absolute position is therefore available only through explicitly controlled filter-state features, making it possible to separate local event evidence from global motion-prior information.

\subsection{Metrics}

Adhering to the evaluation protocol of the FRED dataset, the predictions are evaluated against future boxes using the Average Displacement Error
\begin{align}
  \mathrm{ADE}_\text{center} &= \frac{1}{T}\sum_{k=1}^{T}
  \left\|\hat{\mathbf{c}}_{t_k} - \mathbf{c}_{t_k}\right\|_2,
\end{align}
and the Final Displacement Error
\begin{align}
  \mathrm{FDE}_\text{center} &= 
  \left\|\hat{\mathbf{c}}_{t_T} - \mathbf{c}_{t_T}\right\|_2,
\end{align}
reported in pixels.
Here $\mathbf{c}_{t} = [x_{t}, y_{t}]$ and $\hat{\mathbf{c}}$ is the box center, and its estimate, respectively, at time $t$.
The time $t_T$ refers to the time of the last future prediction.
We also report bounding-box ADE/FDE in pixels and mean IoU over the forecast horizon.
Center metrics measure trajectory accuracy, whereas box metrics and mIoU additionally capture changes in apparent target size.

\subsection{Training and Test Split Policy}

We use two different splits of the FRED dataset for evaluation.
The first is the original, canonical split, and the second is a decorrelated subset.

For the canonical-split experiments, models are trained on the FRED training split and evaluated on the canonical FRED test split. During development, we reserve 20\% of the training recordings at the folder level for checkpoint selection and training evaluation. This avoids mixing windows from the same recording across training and validation. To keep experiments comparable and computationally manageable, training samples are capped at 80k and train-evaluation samples at 20k unless otherwise stated. 
The selected checkpoint is then evaluated on the full test split.

The decorrelated experiments use the same train/test protocol, but apply the sample-level decorrelation procedure introduced in \cref{sec:greedy_decorrelation} separately to the relevant sample pools.
First we select 80k samples from the canonical training split, and decorrelation then removes 50\% of these, resulting in 40k samples.
A second set of 40k samples are then sampled uniformly from the first 80k samples which gives a dataset of the same size but with the same statistics as the canonical dataset split.

\section{Method}
This section describes our residual Kalman filter, two baselines that we compare against, and the decorrelation procedure required for motion prior diagnostics.

\subsection{Optimized Kalman Baselines}

We use a constant-velocity Kalman filter as the physical forecasting baseline and as the starting point for the residual models. The filter operates on the full center-size box state in \cref{eq:kalman_state}
where the first four components, $[x_t,y_t,w_t,h_t]$ are the normalized bounding-box coordinates and the last four, $[v_t^x, v_t^y, v_t^w, v_t^h]$, are their corresponding velocities. This can be viewed as four independent constant-velocity models, one for each box channel ($x,y,w,h$).

The filter directly observes the normalized box position and size, while the corresponding velocities are inferred from the observed history.

The filter is initialized fresh for each sample, filters only the observed history window, and then propagates forward without using future measurements. 
Initial, process, and measurement standard deviations are optimized on the training split. 
The primary optimized baseline uses differentiable backpropagation through the filter with log-space standard-deviation parameters and a weighted objective
\begin{equation}
  J = \mathrm{FDE}_\text{center} + \lambda_\text{ade}\mathrm{ADE}_\text{center}
      - \lambda_\text{iou}\mathrm{mIoU},
\end{equation}

where we used $\lambda_\text{ade}=0.25$ and $\lambda_\text{iou}=100$ during training.
The resulting baseline is therefore not intended as a weak reference model, but as a carefully tuned constant-velocity prior. This is important for the residual formulation: learned models should improve over a competitive motion model rather than over an under-optimized extrapolator.

As an additional baseline, we also optimize an otherwise matched constant-acceleration Kalman filter.

\subsection{Last-Four Linear Extrapolator}

For comparison, we also report a linear extrapolator, intended to replicate the one used by \HariPrasanth but extended to support full bounding boxes.
It estimates velocity for each box channel by a least-squares line fit over the last four observed boxes and then propagates forward with constant velocity:
\begin{equation}
  \hat{\bbox}_{t_k} = \bbox_{t_0} + \hat{\mathbf{v}}(t_k - t_0).
\end{equation}

This baseline is very simple, but useful for interpreting whether the optimized Kalman filter mainly improves center extrapolation, box-size extrapolation, or both.

\subsection{Residual Kalman Forecaster}
\label{sec:residual_kalman}

The learned model predicts acceleration residuals rather than boxes directly. 
Starting at the final filtered Kalman state the residual head predicts the acceleration at forecast step $k$,
\begin{equation}
  \mathbf{a}_k = f_\theta(\phi_\repr, \phi_h, \state_k, \Delta t_k),
\end{equation}
where $\phi_\repr$ is the event representation feature and $\phi_h$ encodes recent box history. The rollout is
\begin{align}
  \bbox_{k+1} &= \bbox_k + \mathbf{v}_k \Delta t_k
    + \frac{1}{2}\mathbf{a}_k \Delta t_k^2, \\
  \mathbf{v}_{k+1} &= \mathbf{v}_k + \mathbf{a}_k \Delta t_k.
\end{align}
The predicted box channels are clamped to $[0,1]$.

This formulation separates a stable physical prior from learned residual structure. The Kalman filter explains the dominant constant-velocity component, while the learned model is responsible only for the remaining acceleration-like correction. This also makes the diagnostic analysis more direct: if future acceleration can be predicted from position and velocity alone, then part of the residual task can be solved without event evidence.

\subsection{Network Structure}

The image encoder is a ResNet-18-style convolutional backbone with three input channels and a 128-dimensional pooled output. The history encoder receives the final $N_h$=12 history samples. Each sample contributes four normalized box values and one relative timestamp, giving a $12 \times 5$ input. In the leakage-reduced setting, box positions are expressed relative to the final history box.
The box input sequence is flattened and mapped by an MLP to a fixed-dimensional history feature $\phi_h$.

The residual head, which is a two-layer MLP with hidden dimension 256, receives the concatenation of the image feature, history feature, current 8-D Kalman rollout state, and scalar time step. It outputs four acceleration channels corresponding to the center and size components of the box. Models without image input use the same residual-rollout structure but omit the image feature, allowing us to isolate the contribution of event-derived evidence.

\subsection{Motion-Prior Residual Baselines and Correlation Analysis}
To estimate how much of the residual forecasting problem can be solved without image evidence and box history, we include a linear residual baseline that uses the same acceleration-rollout formulation as the learned residual forecaster, but replaces the neural residual head with a single linear map from filtered center position and velocity to acceleration. Strong performance from this baseline indicates that part of the future acceleration signal is predictable from simple motion-prior features rather than from event evidence.

We further quantify this effect by fitting a constant-acceleration curve to the center positions over each history-plus-future sample window,
\begin{equation}
\mathbf{c}(t) \approx \mathbf{c}_0 + \mathbf{v}_0(t-t_0) + \frac{1}{2}\mathbf{a}(t-t_0)^2.
\end{equation}
This yields an anchor position $\mathbf{c}_0$, anchor velocity $\mathbf{v}0$, and fitted future acceleration $\mathbf{a}=[a_x,a_y]^\top$. We then measure how predictable $\mathbf{a}$ is from motion-prior features,
\begin{equation}
\mathbf{x}_{prior} =
[c^x_0-0.5, c^y_0-0.5, v^x_0, v^y_0,
|\mathbf{c}_0-[0.5,0.5]|_2, |\mathbf{v}_0|_2].
\end{equation}

We report mean absolute Pearson correlation between these prior features and fitted acceleration, as well as ridge-linear $R^2$. High predictability indicates that the residual target contains structure that can be explained from anchor position and velocity alone. Such structure may be useful for in-distribution forecasting, but it can also mask whether event-conditioned models are exploiting visual evidence.

\subsection{Greedy Decorrelation}
\label{sec:greedy_decorrelation}

The decorrelation method drops samples. For a candidate subset $S$, sufficient statistics are accumulated for standardized prior features $X_S$ and fitted accelerations $Y_S$. 
The score combines mean absolute Pearson correlation and ridge-linear predictability:
\begin{equation}
  \mathrm{score}(S) =
  \alpha \cdot \mathrm{mean}\left(|\mathrm{corr}(X_S,Y_S)|\right)
  + \beta \cdot \mathrm{mean}\left(\max(R^2,0)\right)
\end{equation}

At each step, the algorithm samples a set of removable candidates and removes the example whose removal most reduces the score. The process stops when the target fraction is reached. In our decorrelated experiments, we keep 50\% of the samples.

The procedure removes only the measured linear relationship between selected prior features and fitted acceleration, and it does so by dropping samples rather than by reweighting the data-generating process. 
Nonlinear priors, unmeasured scene statistics, and other forms of shortcut information may remain. We therefore use the decorrelated subsets as a stress test: a model whose gains persist after decorrelation is less dependent on these measured position- and velocity-based shortcuts.

\section{Experiments}

\subsection{Protocol}

All learned models use the optimized Kalman filter as the starting state for forecasting. We compare the Kalman baselines, the last-four linear extrapolator, linear residual baselines, non-image residual models, and image-conditioned residual models. Unless otherwise stated, models use a 400 ms history window, either a 400 ms or 800 ms forecast horizon, normalized boxes, AdamW optimization with learning rate $10^{-3}$, batch size 128, weight decay $10^{-4}$, Smooth L1 loss, and gradient clipping at norm 10.

The main image-conditioned experiments use fixed target-centered cutouts, relative box-history features, filtered state features, and ResNet-18 encoders. We evaluate event frames and CSTR to compare two event-derived representations, and we include non-image variants to separate the contribution of event evidence from the contribution of box history and Kalman state features.

\subsection{Canonical Split Results}

\begin{table*}[hbtp]
\centering
\resizebox{\textwidth}{!}{%
\begin{tabular}{lcccccccc}
\toprule
& & \multicolumn{3}{c}{Short-term, 0.4s} & \multicolumn{3}{c}{Mid-term, 0.8s} \\
Method & Reference & Inputs & (ADE/FDE)$_{BB}$ & (ADE/FDE)$_{C}$ & mIoU & (ADE/FDE)$_{BB}$ & (ADE/FDE)$_{C}$ & mIoU \\
\midrule
CV Opt. Kalman & Our & Boxes & 22.24/39.89 & 19.12/37.25 & 0.475 & 45.81/96.80 & 43.10/94.67 & 0.315 \\
CA Opt. Kalman & Our & Boxes & 22.07/39.37 & 18.91/36.69 & 0.480 & 45.25/95.61 & 42.52/93.51 & 0.316 \\
CV Lin. Approx. & Our & Last 4 Boxes & 31.00/54.35 & 20.43/39.38 & 0.408 & 59.32/117.81 & 45.13/97.80 & 0.264 \\
\midrule
Lin. Regr.$\to$Acc & Our & FS Center Pos+Vel & 21.14/37.00 & 17.96/34.26 & 0.468 & 41.47/84.16 & 38.70/81.93 & 0.301 \\
MLP$\to$Acc & Our & Boxes+FS & 18.20/30.96 & 14.94/28.08 & 0.499 & 34.77/70.47 & 31.89/68.06 & 0.336 \\
CNN$\to$Acc & Our & EF+Boxes+FS & 17.06/28.31 & 13.61/25.21 & 0.522 & \textbf{31.75}/\textbf{64.61} & \textbf{28.62}/\textbf{61.96} & \textbf{0.361} \\
CNN$\to$Acc & Our & CSTR+Boxes+FS & \textbf{16.70}/\textbf{27.92} & \textbf{13.22}/\textbf{24.72} & \textbf{0.531} & 31.92/65.03 & 28.84/62.50 & 0.356 \\
\midrule

CNN+Transformer & \cite{Magrini2025-FRED} & EF+Boxes & 48.46/66.67 & 124.00/45.92 \footnotemark & 0.284 & 72.85/119.60 & 280.90/83.86 \footnotemark[\value{footnote}] & 0.206 \\
CNN+Transformer & \cite{Magrini2025-FRED} & EF+RGB+Boxes & 47.49/65.89 & 121.30/45.23 \footnotemark[\value{footnote}] & 0.279 & 72.81/122.50 & 281.50/85.78 \footnotemark[\value{footnote}] & 0.201 \\
CV Kalman & \cite{HariPrasanth2026} & Box Centers+RPM & -/- & 15.35/34.85 & - & -/- & 31.16/77.76 & - \\

\midrule
\bottomrule
\end{tabular}
}
\caption{Canonical FRED split results. Proposed methods use the optimized full-box constant-velocity Kalman filter as the forecasting prior and, when applicable, learn acceleration residuals on top of it. Metrics are reported for 400 ms and 800 ms horizons; lower ADE/FDE and higher mIoU are better. Best values are highlighted. Comparisons to previously reported methods are included for context but differ in evaluation details or training splits. FS and EF refers to Kalman filter state, and event frames, respectively.}
\label{tab:standard_no_decorr}
\end{table*}

Table \ref{tab:standard_no_decorr} reports results on the canonical FRED split. The optimized constant-velocity and constant-acceleration Kalman filters are strong baselines, especially for bounding-box prediction. The constant-acceleration filter performs slightly better across all reported metrics. Compared with the last-four linear extrapolator, the Kalman filters give similar center performance but substantially better bounding-box ADE/FDE and mIoU, indicating that filtering the full center-size state is important for stable box forecasts.

Residual acceleration models improve over the Kalman baselines across both forecast horizons. The linear residual model using filtered center position and velocity already improves center accuracy, showing that some future acceleration is predictable from simple non-image motion features. The MLP residual model using boxes and filter-state features improves further, and adding event-derived image representations gives the best results. This supports the residual formulation: the constant-velocity Kalman filter captures the dominant constant-velocity component, while learned residuals capture systematic acceleration corrections.

The difference between event frames and CSTR is small on the canonical split. CSTR gives the best short-horizon results, while the two event representations are close at the mid horizon. This suggests that, under the canonical evaluation, the main gain comes from combining residual learning with event-conditioned local evidence rather than from a decisive advantage of one event representation.

The comparison to previously reported FRED and RPM-modulated Kalman results should be interpreted cautiously because of differences in evaluation details, splits, and forecast targets. Nevertheless, the results show that a carefully optimized full-box Kalman baseline plus learned acceleration residuals is highly competitive and substantially stronger than direct learned forecasting baselines reported for FRED.

\footnotetext{Values are reproduced as reported by the original authors. Some reported center ADE/FDE entries appear inconsistent with the usual ADE/FDE ordering. We were unable to confirm corrections with the authors before publication.}

\subsection{Motion-Prior Correlation}
\label{sec:motion_prior_corr}

The canonical results show that position- and velocity-only residual models perform surprisingly well. We therefore analyze whether fitted future acceleration is predictable from anchor position and velocity. Table \ref{tab:correlation} reports mean absolute Pearson correlation and ridge-linear $R^2$ from motion-prior features to fitted center acceleration before and after decorrelation using the method in \cref{sec:greedy_decorrelation}.

\begin{figure}[htbp]
    \centering

    \begin{subfigure}{1.0\textwidth}
        \centering
        \includegraphics[width=\linewidth]{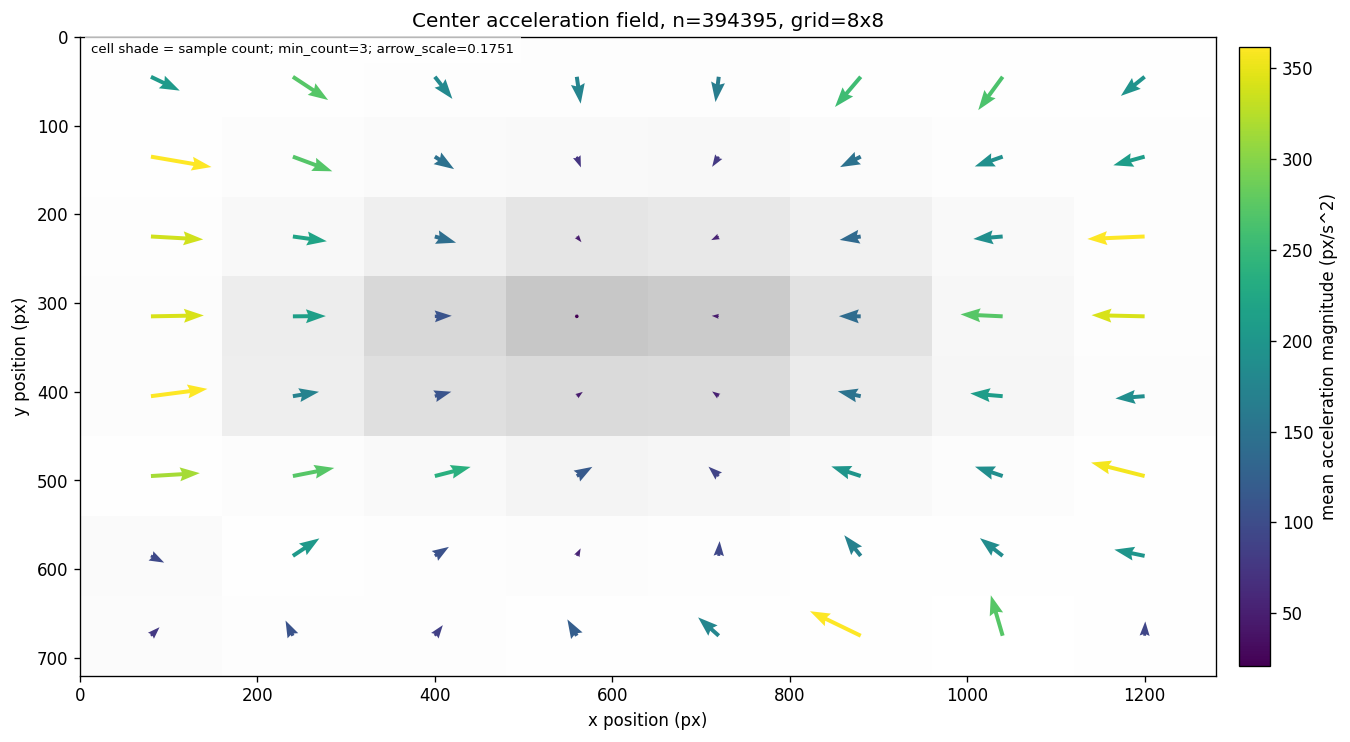}
        \caption{Box center position to acceleration. }
    \end{subfigure}
    \begin{subfigure}{1.0\textwidth}
        \centering
        \includegraphics[width=\linewidth]{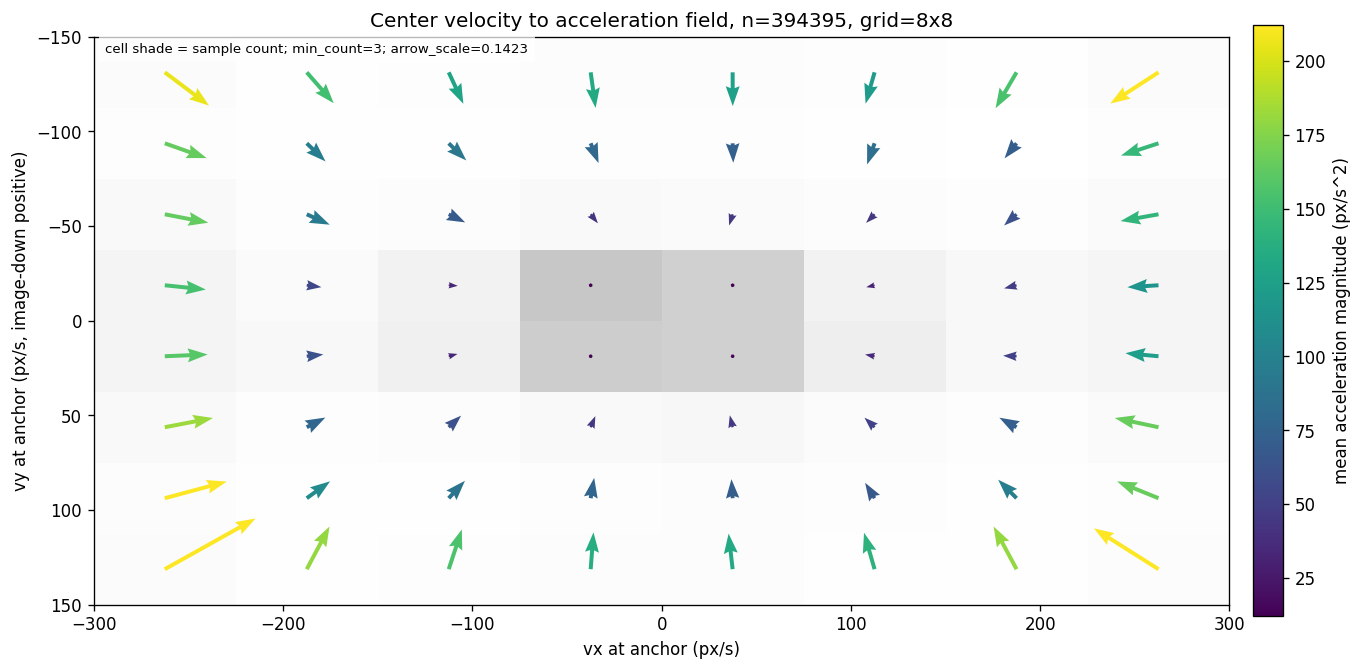}
        \caption{Estimated velocity to acceleration. }
    \end{subfigure}

    \caption{Motion-prior structure in FRED for the canonical train split. Samples are binned by anchor box-center position in (a) and estimated anchor velocity in (b). Arrows show the mean fitted future center acceleration in each bin, with direction encoding acceleration direction and length/color encoding magnitude. Cell shading indicates sample count. The fields reveal edge-to-center and velocity-dependent deceleration patterns, showing that future acceleration is partly predictable from position and velocity alone. }
    \label{fig:correlation_illustrations}
\end{figure}

Before decorrelation, both the training and test splits show non-negligible predictability. This confirms that the residual target is not purely visual: part of the future acceleration field can be explained from where the UAV is in the image and how it is moving at the anchor time. Figure \ref{fig:correlation_illustrations} visualizes this effect. The position-to-acceleration field shows edge-to-center tendencies, while the velocity-to-acceleration field shows velocity-dependent deceleration patterns. These effects are consistent with motion-prior shortcuts: a model may learn that objects near certain image regions or moving at certain speeds tend to accelerate in predictable directions.

After greedy decorrelation, the measured correlations and linear predictability are greatly reduced. This supports the use of the decorrelated subsets as a stricter diagnostic setting. The decorrelated subsets do not remove all possible priors, but they substantially weaken the measured position- and velocity-based shortcuts targeted by the procedure.

\begin{table}[t]
  \centering
  \begin{tabular}{lcc|cc}
    \toprule
    & \multicolumn{2}{c|}{Before} & \multicolumn{2}{c}{After} \\   
    Split & Mean $|\rho|$ & Mean $R^2$ & Mean $|\rho|$ & Mean $R^2$ \\
    \midrule
    Train & 0.128 & 0.244 & 0.008 & 0.003 \\
    Test & 0.131 & 0.252 & 0.007 & 0.003 \\
    \bottomrule
  \end{tabular}
  \caption{Motion-prior predictability of fitted future acceleration before and after sample decorrelation.}
  \label{tab:correlation}
\end{table}

\subsection{Decorrelated-Subset Results}

Tables \ref{tab:compare_decorr_orig_train_orig_test} and \ref{tab:compare_orig_decorr_train_decorr_test} evaluate how models behave when the motion-prior shortcuts measured in Section \ref{sec:motion_prior_corr} are weakened. Table \ref{tab:compare_decorr_orig_train_orig_test} shows results when models are trained on decorrelated training samples and evaluated on the canonical test split. Performance generally degrades relative to training on the full, canonical, training split, which is expected: the decorrelated training data removes part of the structure that remains present in the canonical test data. This result should not be interpreted as a failure of decorrelation, but as evidence that the canonical split contains motion-prior regularities that are useful when evaluated on the canonical test split.

\begin{table*}[bt]
\centering
\resizebox{\textwidth}{!}{%
\begin{tabular}{llccccc}
\toprule
Horizon & Inputs & ADE$_{BB}$ & FDE$_{BB}$ & ADE$_{C}$ & FDE$_{C}$ & mIoU \\
\midrule
Short (400 ms) & FS Center Pos+Vel & 21.50 (+1.7\%) & 37.92 (+2.5\%) & 18.28 (+1.8\%) & 35.03 (+2.2\%) & 0.475 (+1.5\%) \\
 & Boxes+FS & 19.15 (+5.2\%) & 33.50 (+8.2\%) & 15.92 (+6.6\%) & 30.59 (+8.9\%) & 0.485 (-2.8\%) \\
 & EF+Boxes+FS & 17.30 (+1.4\%) & 29.42 (+3.9\%) & 13.81 (+1.5\%) & 26.23 (+4.0\%) & 0.523 (+0.2\%) \\
 & CSTR+Boxes+FS & \textbf{17.00} (+1.8\%) & \textbf{28.67} (+2.7\%) & \textbf{13.51} (+2.2\%) & \textbf{25.44} (+2.9\%) & \textbf{0.529} (-0.4\%) \\
\midrule
Mid (800 ms) & FS Center Pos+Vel & 43.20 (+4.2\%) & 89.32 (+6.1\%) & 40.36 (+4.3\%) & 86.90 (+6.1\%) & 0.320 (+6.3\%) \\
 & Boxes+FS & 38.89 (+11.8\%) & 83.20 (+18.1\%) & 36.04 (+13.0\%) & 80.73 (+18.6\%) & 0.321 (-4.5\%) \\
 & EF+Boxes+FS & 33.50 (+5.5\%) & 70.21 (+8.7\%) & 30.35 (+6.0\%) & 67.44 (+8.8\%) & 0.356 (-1.4\%) \\
 & CSTR+Boxes+FS & \textbf{33.31} (+4.4\%) & \textbf{70.07} (+7.8\%) & \textbf{30.16} (+4.6\%) & \textbf{67.40} (+7.8\%) & \textbf{0.358} (+0.6\%) \\
\bottomrule
\end{tabular}
}
\caption{Effect of training-set decorrelation when evaluating on the canonical FRED test split. All models are trained on 50\% of the available training samples. The decorrelated setting uses the greedy decorrelation procedure, while the reference setting uses a random 50\% subset. Percentages show relative change with respect to the corresponding model trained on the random subset. Positive values indicate higher error for ADE/FDE and lower IoU for mIoU.
FS and EF refer to Kalman filter state, and event frames, respectively.}
\label{tab:compare_decorr_orig_train_orig_test}
\end{table*}

Table \ref{tab:compare_orig_decorr_train_decorr_test} instead shows results when models are trained on the canonical training split and evaluated on the decorrelated test subset. In this setting, models trained on the canonical data often perform worse than models trained on decorrelated data, especially for the 800 ms horizon. This suggests a train-test mismatch: models trained on the canonical split can benefit from motion-prior correlations that are weakened in the decorrelated test set. Training on decorrelated data reduces this dependence and can improve relative robustness under the diagnostic stress test.

Across decorrelated settings, event-conditioned models remain stronger than non-image residual models. The gap is not eliminated by decorrelation, indicating that local event evidence and box history still provide useful information beyond the measured position- and velocity-based priors. CSTR seems to be more stable than event frames and consistently yields better results under non-matching train/test decorrelation conditions, although the difference is modest. Overall, the decorrelated experiments support the central claim: canonical-split gains should be interpreted cautiously, but residual event-conditioned models retain useful predictive signal even when simple motion-prior shortcuts are reduced.

\begin{table*}[bt]
\centering
\resizebox{\textwidth}{!}{%
\begin{tabular}{llccccc}
\toprule
Horizon & Inputs & ADE$_{BB}$ & FDE$_{BB}$ & ADE$_{C}$ & FDE$_{C}$ & mIoU \\
\midrule
Short (400 ms) & FS Center Pos+Vel & 24.64 (+3.1\%) & 43.76 (+4.7\%) & 21.40 (+4.3\%) & 41.03 (+5.9\%) & 0.434 (-5.4\%) \\
 & Boxes+FS & 21.01 (+0.8\%) & 36.08 (+0.8\%) & 17.62 (+1.1\%) & 33.08 (+1.2\%) & 0.473 (-0.8\%) \\
 & EF+Boxes+FS & 19.33 (-1.0\%) & 32.38 (-1.3\%) & 15.73 (-0.9\%) & 29.16 (-0.7\%) & 0.502 (+0.6\%) \\
 & CSTR+Boxes+FS & \textbf{19.15} (-0.3\%) & \textbf{32.07} (-0.1\%) & \textbf{15.49} (-0.4\%) & \textbf{28.77} (+0.1\%) & \textbf{0.507} (+0.0\%) \\
\midrule
Mid (800 ms) & FS Center Pos+Vel & 45.23 (+5.8\%) & 91.55 (+7.1\%) & 42.56 (+6.9\%) & 89.50 (+8.0\%) & 0.283 (-12.4\%) \\
 & Boxes+FS & 37.99 (+2.4\%) & 76.92 (+1.4\%) & 35.16 (+2.9\%) & 74.56 (+1.9\%) & 0.322 (-4.5\%) \\
 & EF+Boxes+FS & 33.91 (+1.4\%) & 68.97 (+2.7\%) & 30.85 (+1.8\%) & 66.44 (+3.2\%) & 0.353 (-2.2\%) \\
 & CSTR+Boxes+FS & \textbf{33.83} (+1.4\%) & \textbf{68.51} (+0.5\%) & \textbf{30.72} (+1.7\%) & \textbf{65.87} (+0.6\%) & \textbf{0.355} (-2.2\%) \\
\bottomrule
\end{tabular}
}
\caption{Effect of test-set decorrelation when evaluating on the decorrelated FRED test subset. Models trained on the canonical training subset are compared with the corresponding models trained on the decorrelated training subset. Percentages show relative change with respect to evaluation on the canonical test subset for the same model and horizon. Positive values indicate higher error for ADE/FDE and lower IoU for mIoU.
 FS and EF refer to Kalman filter state, and event frames, respectively.}
\label{tab:compare_orig_decorr_train_decorr_test}
\end{table*}

\section{Discussion}

The experiments highlight both the strength and the risk of residual forecasting on FRED. 
The CV and CA Kalman filters over the full center-size state are strong, interpretable baselines, and learned acceleration residuals consistently improve over them. 
This supports the idea that short-horizon UAV forecasting benefits from combining a stable physical prior with learned corrections. 

At the same time, the linear residual baselines and correlation analysis show that future acceleration is partially predictable from anchor position and velocity, meaning that canonical-split improvements may reflect both event-conditioned evidence and dataset-specific motion priors.

The decorrelation procedure is therefore best understood as a diagnostic stress test rather than a new benchmark or evidence of out-of-distribution robustness. It weakens selected linear relationships between motion-prior features and fitted acceleration by removing samples, but it does not change the underlying data collection process or eliminate nonlinear priors, scene-dependent cues, target-scale effects, or other shortcuts. Within these limitations, the stress test helps separate gains that depend strongly on measured position- and velocity-based shortcuts from gains that persist when those shortcuts are reduced.

Even so, the stress test is informative. Event-conditioned residual models still perform well after decorrelation, while models relying more directly on position and velocity are more affected. This suggests that event evidence contributes useful information, but also that part of the canonical-split improvement comes from motion-prior structure. Reporting both canonical and decorrelated-subset results gives a more cautious and transparent picture of model behavior.

There are several limitations. The experiments are restricted to short- and mid-horizon box forecasting on FRED, and the decorrelation analysis focuses on fitted center acceleration rather than full box dynamics. 
The residual model produces a single deterministic trajectory and does not model forecast uncertainty, such as multiple distinct plausible maneuvers. Finally, the event representations are evaluated through relatively compact encoders and target-centered cutouts; more specialized event architectures may extract stronger visual evidence.

\section{Conclusion}
\label{sec:conclusion}

We presented a residual Kalman approach for short-horizon UAV bounding-box forecasting that achieves strong results on FRED and improves over our optimized full-box Kalman baselines, while comparing favorably to previously reported methods under non-identical protocols.

The method extends center-only trajectory forecasting to a full center-size Kalman state and learns acceleration-like residuals from event representations, box history, and filtered state features. This combination gives a strong forecasting model: the Kalman filter provides a stable first-order physical prior, while the residual model learns data-dependent acceleration and turning corrections beyond fixed parametric dynamics.

We also introduced a diagnostic analysis for motion-prior dependence. By measuring how well fitted future acceleration can be predicted from anchor position and velocity, we showed that part of the residual forecasting target is explainable without event evidence. A greedy sample decorrelation procedure reduces this predictability and provides a stricter stress test for learned residual models. The resulting experiments suggest that event-conditioned residuals remain useful when measured position- and velocity-based shortcuts are weakened, but they also show that canonical-split gains should be interpreted with caution.

\bibliographystyle{splncs04}
\bibliography{main}

\begin{thebibliography}{10}
\providecommand{\url}[1]{\texttt{#1}}
\providecommand{\urlprefix}{URL }
\providecommand{\doi}[1]{https://doi.org/#1}

\bibitem{bbc2019}
Heathrow airport: Drone sighting halts departures. {BBC}  (2019),
  \url{https://bbc.com/news/uk-46803713}

\bibitem{ModUA2026}
{“Army of Drones Bonus”} program delivers results: nearly 820,000 russian
  targets hit in 2025, says {Mykhailo Fedorov}. Ministry of Defence Ukraine
  (2026),
  \url{https://mod.gov.ua/en/news/army-of-drones-bonus-program-delivers-results-nearly-820-000-russian-targets-hit-in-2025-says-mykhailo-fedorov}

\bibitem{Becker2021}
Becker, S., Hug, R., Huebner, W., Arens, M., Morris, B.T.: Generating synthetic
  training data for deep learning-based uav trajectory prediction. In:
  Proceedings of the 2nd International Conference on Robotics, Computer Vision
  and Intelligent Systems - ROBOVIS. pp. 13--21. INSTICC, SciTePress (2021).
  \doi{10.5220/0010621400003061}

\bibitem{cstr}
El~Shair, Z.A., Hassani, A., Rawashdeh, S.A.: {CSTR: A Compact Spatio-Temporal
  Representation for Event-Based Vision}. IEEE Access  \textbf{11},
  102899--102916 (2023). \doi{10.1109/ACCESS.2023.3316143}

\bibitem{Gallego2022}
Gallego, G., Delbruck, T., Orchard, G., Bartolozzi, C., Taba, B., Censi, A.,
  Leutenegger, S., Davison, A.J., Conradt, J., Daniilidis, K., Scaramuzza, D.:
  {Event-Based Vision: A Survey}. IEEE transactions on pattern analysis and
  machine intelligence  \textbf{44},  154--180 (2022).
  \doi{10.1109/TPAMI.2020.3008413},
  \url{https://ieeexplore.ieee.org/document/9138762/}

\bibitem{Liang2025}
Liang, H., Yuan, S., Liu, F., Yang, Y., Wang, B., Huang, Z., Shi, C., Jin, J.:
  Label-free long-horizon 3d uav trajectory prediction via motion-aligned rgb
  and event cues (2025), \url{https://arxiv.org/abs/2507.03365}

\bibitem{HariPrasanth2026}
M., H.P.S., Habibiroudkenar, P., Alamikkotervo, E., Bouzoulas, D., Ojala, R.:
  {Event-only drone trajectory forecasting with RPM-modulated Kalman
  filtering}. arXiv [cs.CV]  (2026). \doi{10.48550/arXiv.2603.01997},
  \url{http://dx.doi.org/10.48550/arXiv.2603.01997}

\bibitem{Magrini2025-DroneDet}
Magrini, G., Berlincioni, L., Becattini, F., Cultrera, L., Pala, P.: {Drone
  detection with event cameras}. In: {2025 IEEE/CVF International Conference on
  Computer Vision Workshops (ICCVW)}. pp. 4762--4773. IEEE (2025).
  \doi{10.1109/iccvw69036.2025.00494},
  \url{http://dx.doi.org/10.1109/iccvw69036.2025.00494}

\bibitem{Magrini2025-FRED}
Magrini, G., Marini, N., Becattini, F., Berlincioni, L., Biondi, N., Pala, P.,
  Del~Bimbo, A.: {FRED}: The {Florence} {RGB}-event drone dataset. In:
  Proceedings of the 33rd ACM International conference on multimedia (2025)

\bibitem{sivakumar2021uavsurvey}
Sivakumar, M., TYJ, N.M.: A literature survey of unmanned aerial vehicle usage
  for civil applications. Journal of Aerospace Technology and Management
  \textbf{13}, ~1--8 (2021). \doi{10.1590/jatm.v13.1233}

\bibitem{Spetlik2025}
Spetlik, R., Uhrová, T., Matas, J.: {Efficient real-time quadcopter propeller
  detection and attribute estimation with high-resolution event camera}, pp.
  217--230. Lecture notes in computer science, Springer Nature Switzerland
  (2025). \doi{10.1007/978-3-031-95911-0_16},
  \url{http://dx.doi.org/10.1007/978-3-031-95911-0_16}

\bibitem{Stewart2023}
Stewart, T., Drouin, M.A., Picard, M., Billy~Djupkep, F., Orth, A., Gagné, G.:
  {Using neuromorphic cameras to track quadcopters}. In: {Proceedings of the
  2023 International Conference on Neuromorphic Systems}. pp.~1--5. ACM (2023).
  \doi{10.1145/3589737.3605987},
  \url{http://dx.doi.org/10.1145/3589737.3605987}

\bibitem{bbc2025}
Wilson, T.: Copenhagen and oslo airports forced to close temporarily due to
  drone sightings. {BBC}  (2025),
  \url{https://bbc.com/news/articles/cn4lj1yvgvgo}

\bibitem{Zhou2024}
Zhou, S., Yang, L., Liu, X., Wang, L.: Learning short-term spatial–temporal
  dependency for uav 2-d trajectory forecasting. IEEE Sensors Journal
  \textbf{24}(22),  38256--38269 (2024). \doi{10.1109/JSEN.2024.3466516}

\bibitem{Zou2025}
Zou, L., Wang, J., Liang, R., Wu, H., Chen, K., Wang, Y.: {UAV-MM3D}: A
  large-scale synthetic benchmark for 3d perception of unmanned aerial vehicles
  with multi-modal data (2025), \url{https://arxiv.org/abs/2511.22404}

\end{thebibliography}

\end{document}